# Comparison of the language networks from literature and blogs

Sabina Šišović, Sanda Martinčić-Ipšić, Ana Meštrović
Department of Informatics University of Rijeka, Rijeka, Croatia
ssisovic@uniri.hr, smarti@uniri.hr, amestrovic@inf.uniri.hr

**Abstract** - In this paper we present the comparison of the linguistic networks from literature and blog texts. The linguistic networks are constructed from texts as directed and weighted co-occurrence networks of words. Words are nodes and links are established between two nodes if they are directly co-occurring within the sentence. The comparison of the networks structure is performed at global level (network) in terms of: average node degree, average shortest path length, diameter, clustering coefficient, density and number of components. Furthermore, we perform analysis on the local level (node) by comparing the rank plots of in and out degree, strength and selectivity. The selectivity-based results point out that there are differences between the structure of the networks constructed from literature and blogs.

I. INTRODUCTION

The representation and analysis of written texts in terms of graphs and complex networks offers an alternative approach for studying the language with different applications in the domain of natural language processing (NLP). Various types of linguistic networks have already been studied: syntax networks [1,2], semantic networks [3], phonological networks [4], syllable networks [5,6], word co-occurrence networks [7-19]. In [3,20,21] a systematic methodological overview of linguistic complex networks principles is presented. Recently, linguistic co-occurrence networks have been intensively studied in order to analyse the structure of the language [7-19].

As the networks incorporate associations between words and concepts, their structure, quantified by global and local network measures [22], such as clustering coefficient, shortest path, diameter, density, node degree, can provide information on some properties of the text. The motivation of our research was to find which network measures are sensitive on different texts categories.

In our previous research [8, 9] we showed the advantages of using a directed and weighted co-occurrence network as the model to capture the structure of a text. In this work we study global and local network measures for the networks constructed from different categories of texts. In particular, at the local level, we applied the node selectivity measure in order to examine if it is sensitive on different styles of texts. Node selectivity is defined as the average weight distribution on the links of the single node [10]. Therefore, in our approach we constructed directed and weighted co-occurrence networks from different texts: 4 books and 4 blogs. We compare global and local network measures for book-blog network pairs.

In the second section we present the overview of related work. In the third section we present key measures of complex networks. In the fourth section the data and network construction techniques are presented. In the fifth section we present the results. In the last section we elaborate on the obtained data and provide concluding remarks.

II. RELATED WORK

Ferrer i Cancho and Solé in [11] first showed that the co-occurrence networks have a small average path length, a high clustering coefficient, and a two-regime power law degree distribution; the network exhibits small-world and scale-free properties. Drogotsev and Mendes [12] used co-occurrence networks to study language as a self-organising network of interacting words. Masucci and Rodgers in [13] investigated the co-occurrence network topology of Orwell's '1984' focusing on the local properties: nearest neighbours and the clustering coefficient. Furthermore, in [10] they introduced the node selectivity measure that can distinguish the difference between normal and randomised text. Liu and Cong [14] constructed co-occurrence networks from text in different languages and used complex network parameters for the classification (hierarchical clustering) of 14 languages, where Croatian was amongst 12 Slavic.

Different applications of linguistic network analysis in NLP includes: evaluation of language complexity [15], automatic summarisation [16] and evaluation of machine translation [17], authorship attribution [18] and text quality analysis [19].

Costa et al. [15] studied the relationship between the topology of network and complexity of the text. They studied texts with different levels of simplification in co-occurrence networks and found that topological regularity correlated negatively with textual complexity. Furthermore, they showed that strength, shortest path, diversity and hierarchical measures can make a distinction between normal text and simplified text. In [16] the authors describe a method that uses complex networks concepts for the summarisation task. In [17] several metrics from complex networks are exploited in order to evaluate the quality of translations. The best distinctions were obtained with the out-degree, in-degree, minimum path and cluster coefficient. In [18] authors investigate the correlation between the properties of networks and author characteristics. It is shown that the networks produced for each author are sensitive to specific features, which indicates that complex networks can capture author characteristics and, therefore, could be used for the authorship identification. In [19] authors investigate the possibility of automated evaluation of text quality using topological measurements extracted from the corresponding complex networks. All the measures are correlated with grades assigned by human experts. It is shown that out-degree, clustering coefficient and deviation from linear dynamics in the network growth are correlated with the text quality.

However, there are no comprehensive studies focused on finding network measures that are sensitive to different text categories. Our work is the first attempt to analyse which network measures can differentiate between literature and blog networks.

### III. THE NETWORK STRUCTURE ANALYSIS

This section contains explanations of network-based measures [22, 10] that are used in our approach for the comparison of two categories of language networks.

Every network has an *N* number of nodes and *K* number of links. Considering the fact that our networks are weighted every link connecting two nodes has an associated weight. The degree of a node *i* is the number of links with which the node is connected, $k_i$. In the case of the directed network, there are two kinds of the degree: the in-degree, $k_i^{in}$ corresponding to the number of incoming links and the out-degree, $k_i^{out}$ equal to the number of outgoing links. The average degree of the network is:

$$<k> = \frac{2K}{N}. \quad (1)$$

For every two connected nodes *i* and *j*, the number of connections lying on the path between them is represented as $d_{ij}$, and so $d_i$ is the average distance of a node *i* from all other nodes, and it's obtained by:

$$d_i = \frac{\sum_j d_{ij}}{N}. \quad (2)$$

For the next two measures, if a network contains more than one component, we consider the largest component. The average shortest path length between every two nodes in network is:

$$L = \frac{1}{N(N-1)} \sum_{i \neq j} d_{ij}, \quad (3)$$

and the maximum distance results in the network diameter, D:

$$D = max_i d_i. \quad (4)$$

The clustering coefficient is a measure which defines the presence of connections between the nearest neighbours of a node. And so, ci (clustering coefficient) of a node is a fraction between the number of edges Ei that exist between that ki and the total possible number:

$$c_i = \frac{2E_i}{k(k-1)}. \quad (5)$$

The average clustering of a network is defined as the average value of the clustering coefficients of all nodes in a network:

$$C = \frac{1}{N} \sum_i c_i. \quad (6)$$

Density of network is a measure of network cohesion defined as the number of observed relationships divided by the number of possible relationships:

$$d = \frac{K}{N(N-1)}. \quad (7)$$

Strength of the node i is the number of its outgoing and incoming links (sum of its weights). For the directed networks the in-strength and the out-strength are defined:

$$s_i^{out/in} = \sum_j w_{ij/ji}. \quad (8)$$

The node selectivity measure can capture the effective distribution of numbers in the weighted adjacency matrix, and it is obtained as a ratio of (out/in-) node strength and its (out/in-) degree:

$$e_i^{out/in} = \frac{s_i^{out/in}}{k_i^{out/in}}. \quad (9)$$

All presented measures are standard network measures usually used for network structure analysis, except the node selectivity measure which is introduced in [10] as the measure that can differentiate between networks based on normal and randomized texts. According to these results, we expected that node selectivity may be potentially important for the text categories differentiation and include it in the set of standard network measures.

In order to illustrate the relationships between node degree, node strength (which are the standard local network measures) and node selectivity, we constructed a small network of seven nodes presented in Figure 1.

Additionally, Table 1 contains values of in-degree, out-degree, in-strength, out-strength and in-selectivity and out-selectivity for all seven nodes in the network presented in Figure 1.

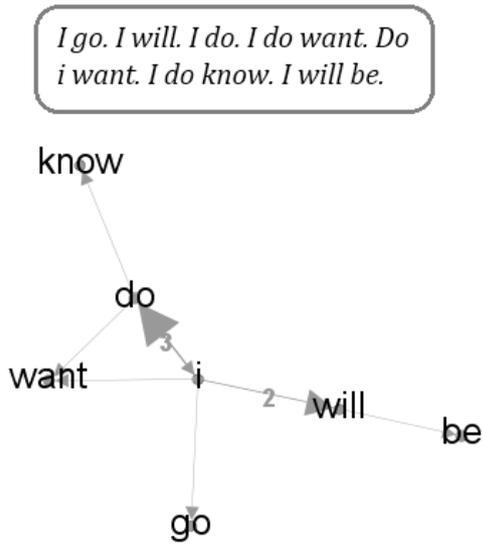

Figure 1  Weighted and directed co-occurrence network of seven nodes, created from the short text below

TABLE 1 - VALUES OF IN/OUT - DEGREE, STRENGTH AND SELECTIVITY

| NODE | $k^{in}$ | $k^{out}$ | $w^{in}$ | $w^{out}$ | $s^{in}$ | $s^{out}$ |
|---|---|---|---|---|---|---|
| i | 1 | 4 | 1 | 7 | 1 | 1,75 |
| do | 1 | 2 | 3 | 2 | 3 | 1 |
| will | 1 | 1 | 2 | 1 | 2 | 1 |
| want | 2 | 0 | 2 | 0 | 1 | 0 |
| know | 1 | 0 | 1 | 0 | 1 | 0 |
| go | 1 | 0 | 1 | 0 | 1 | 0 |
| be | 1 | 0 | 1 | 0 | 1 | 0 |

## IV. NETWORK CONSTRUCTION

### A. Data

Our corpus contains 4 books written or translated into the Croatian language, and 4 blog texts written in Croatian language. The books are: *Picture of Dorian Gray, Bones, The Return of Philip Latinowicz* and *Mama Leone*. The blogs are: *Index.hr, Slobodna Dalmacija, Narodne novine and Gospodarski list* (daily newspaper portal, or business portal). The feature which prompted us to do the comparison is the linguistic distinction between book and blog. Books are written in formal language, standard expressions and phrases are used, whilst blogs are mostly written in a casual mode, with the use of slang, the shortenings of the words or mistakes in syntax. Books come in different sizes and so we compared them with the approximately same sized blog (with the same number of different words), which means we had 4 book-blog pairs for comparison. The sizes of books and blogs in number of total words are shown in the first row of Table 2, while the numbers of different words are presented in the second row (as the number of nodes).

### B. The construction of co-occurrence networks

We used Python and the NetworkX software package developed for the creation, manipulation, and study of the structure, dynamics, and functions of complex networks [23].

The first step in creating networks was text "cleaning": normalising symbols for Croatian diacritics (č, ć, ž, đ, and š), removing special symbols and removing punctuation which does not mark the end of a sentence. We created 8 networks, weighted and directed. Nodes are words that are linked if they are direct neighbours in a sentence. The next step was creating the networks as weighted edgelists, which contain all the pairs of connected words and their weights (the number of connections between two same words).

## V. RESULTS

In this section we present the results of our measuring described in section 2, such as average degree $<k>$, average path distance $L$, diameter $D$, and the average clustering coefficient $C$, density $d$, node strength $s_i$ and node selectivity $e_i$.

In Table 2 we present the estimated global network measures. There are certain differences between measures for book-blog network pairs, but there is no uniform rule that may be used to differentiate these two styles of writing.

TABLE 2 - THE COMPARISON OF NETWORK MEASURES FOR BOOK-BLOG NETWORK PAIRS

| Measure | Text | | | | | | | |
|---|---|---|---|---|---|---|---|---|
| | Bones | Gospodarski list | Mama Leone | Narodne novine | Picture of Dorian Gray | Index.hr | Return of Phillip Latinowicz | Slobodna Dalmacija |
| Number of different words | 191 986 | 199 417 | 85 347 | 146 731 | 75 099 | 118 548 | 28 137 | 44 367 |
| Number of nodes (*N*) | 27396 | 27727 | 13067 | 13036 | 15631 | 15065 | 9531 | 9553 |
| Number of edges (*K*) | 102052 | 105171 | 49383 | 55661 | 46201 | 28972 | 21760 | 25155 |
| Average degree (<*k*>) | 7,45 | 7,58 | 7,56 | 8,54 | 3,88 | 3,85 | 4,57 | 5,27 |
| Avg. shortest path (*L*) | 3,21 | 3,28 | 3,11 | 3,17 | 3,45 | 3,45 | 3,59 | 3,56 |
| Diameter (*D*) | 10 | 21 | 10 | 12 | 14 | 22 | 16 | 13 |
| Average clustering coefficient (*C*) | 0,25 | 0,22 | 0,29 | 0,22 | 0,18 | 0,016 | 0,15 | 0,17 |
| Density (*d*) | 0,0002 | - | 0,00056 | 0,00066 | 0,0004 | 0,0002 | 0,00048 | 0,00055 |
| No. of connected components | 15 | 7 | 1 | 2 | 1 | 45 | 5 | 3 |

Furthermore, we compare networks on the node-level using degree, strength and selectivity measures. For the purpose of comparison we used rank plots. The in/out-degree rank function represents the relationship function between the rank and the in/out-degree of the degree sequence of all nodes sorted in decreasing order.

Similarly, the in/out-strength rank plot and the in/out-selectivity rank plot are defined.

Figure 2 represents the results of the comparisons of the in-degree rank plot and the out-degree rank plot for one book-blog network pair.

The plots do not show significant difference for the in-degree nor for the out-degree rank plots between book-blog network pair. We also experimented with additional three book-blog pairs and we obtained similar results (not reported here due to limited space).

The results of the comparisons of the in/out-strength rank plot for the same book-blog network pair are shown in Figure 3. Again, there is no difference except some small deviation that can be noticed in the plot, but we cannot conclude that in/out-strength distinguish books from blogs.

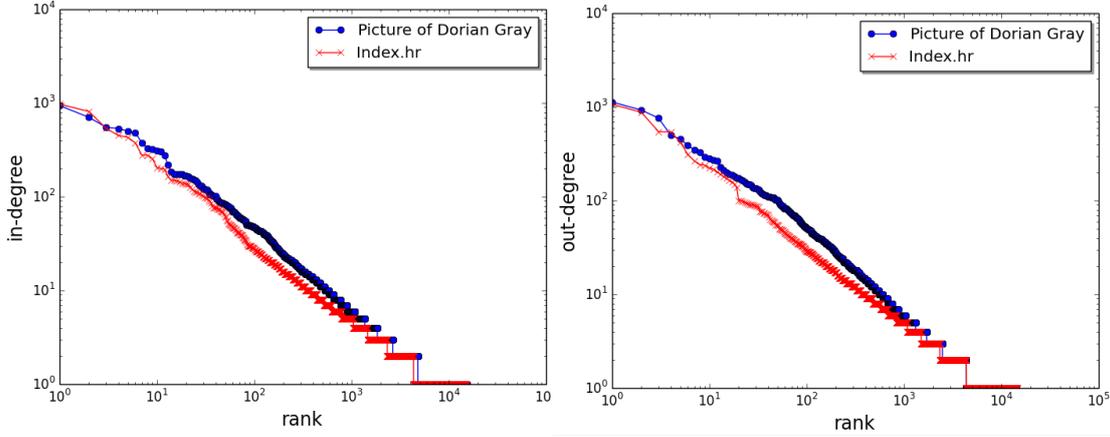

Figure 2  in-degree and out-degree rank plots

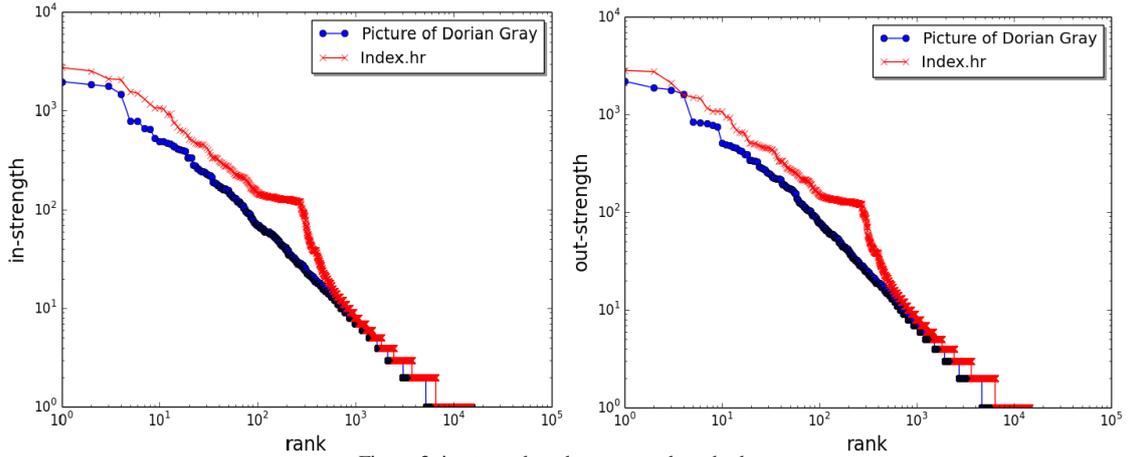

Figure 3  in-strength and out-strength rank plots

The selectivity rank plots are shown in Figure 4 (in-selectivity) and in Figure 5 (out-selectivity). The results show that there are differences in selectivity rank plots between networks constructed from books and networks constructed from blogs for all 4 book-blog network pairs.

In general, all in/out-selectivity values are lower for books than for blogs. We disregarded nodes with zero values of degree because it causes the division by zero (in total 4% of nodes).

## VI.   CONCLUSION

In this work we analysed which complex network measures can distinguish between different texts categories: literature and blogs.

Our results indicate that global network measures are not precise enough to capture the structural differences between networks constructed from different text categories. Even the compared in/out- degree rank plots and in/out- strength rank plots on the local level do not clearly show the differences. However, in-selectivity and out-selectivity rank plots indicate that there are structural

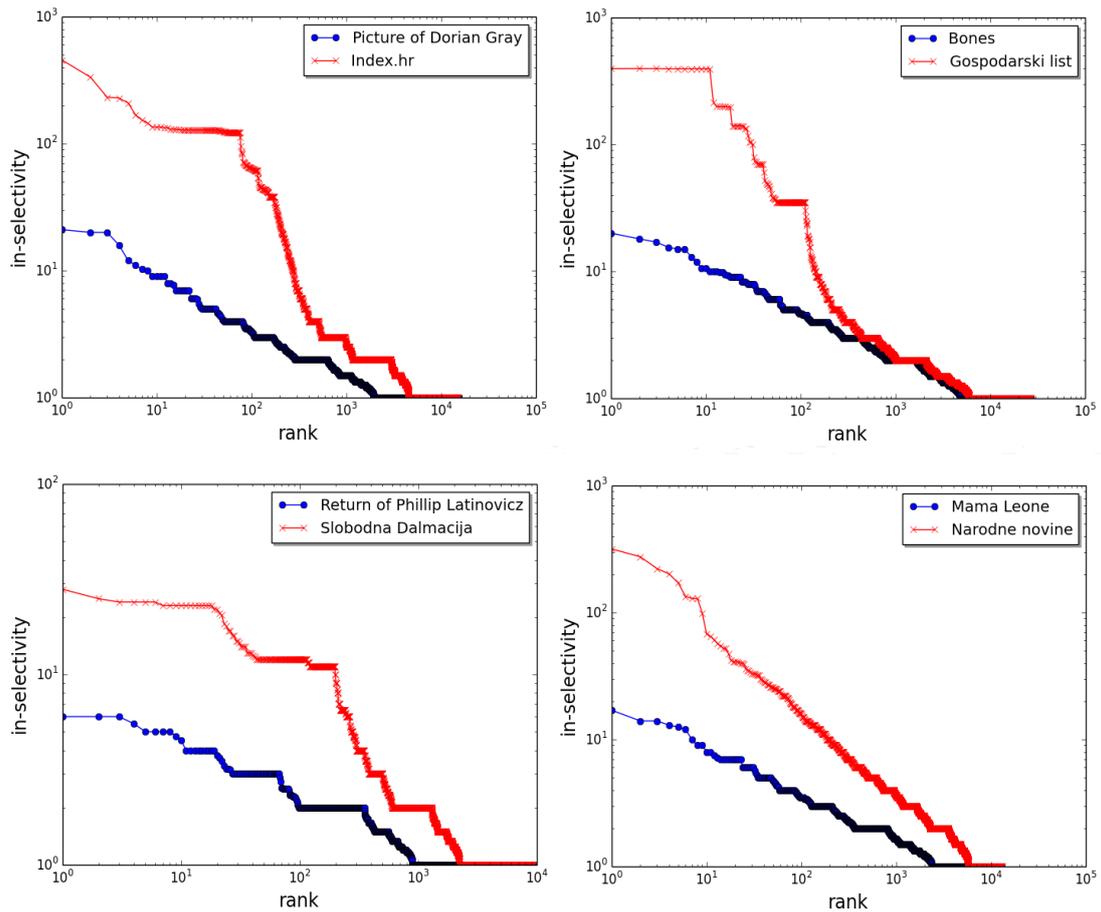

Figure 4 in-selectivity rank plots for 4 book-blog network pairs

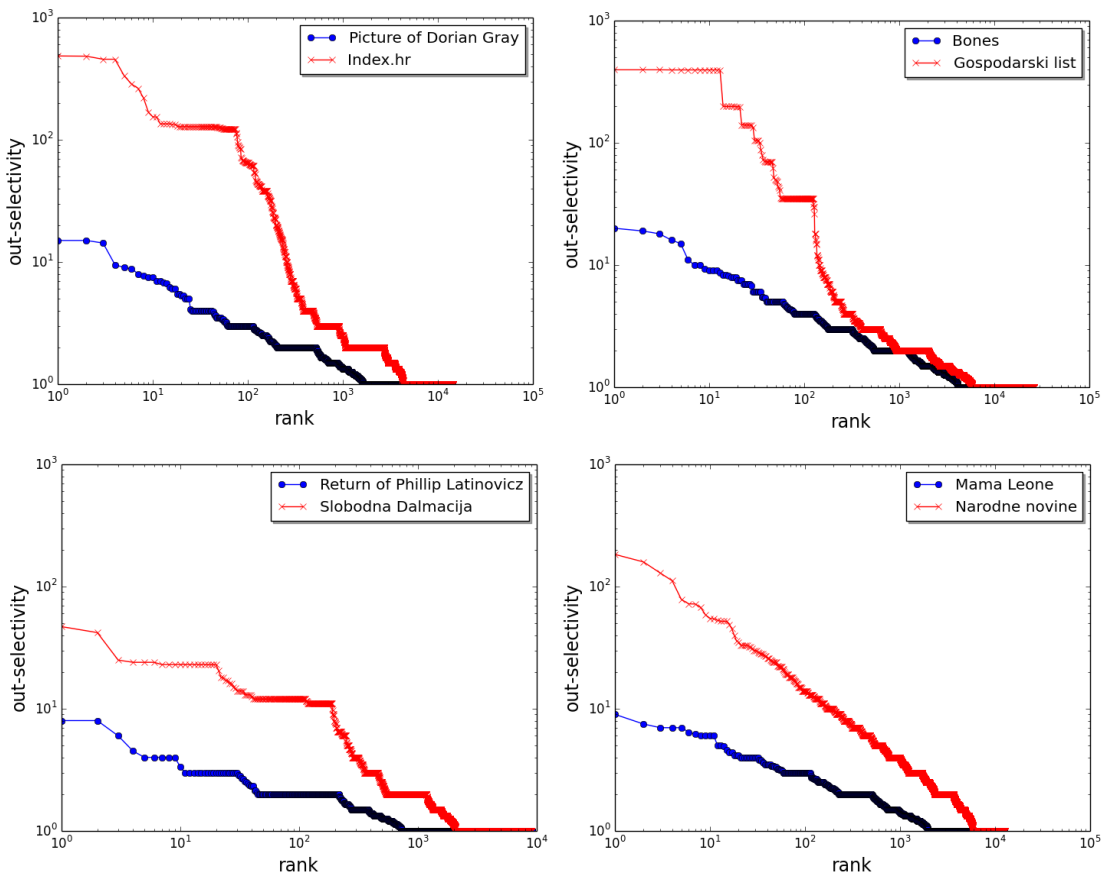

Figure 5 out-selectivity rank plots for 4 book-blog network pairs and networks constructed from blogs. The values of in- and out- node selectivity measure are higher for networks generated from blogs than for networks generated from books. The node selectivity based measures are the only measures from the set of tested measures that can capture the structural differences between two classes of networks.

The presented approach should be extended with examination of other network measures that depend on the strength, degree and link direction in combination with other measures such as clustering coefficient. For future work we will test the broader set of measures on various categories of texts. We plan to include forum comments, texts from Wikipedia, poems and other genres of literature in our corpus. Obtained results encourage us to investigate the complex network properties for text classification, text evaluation or even text quality assessment.

## VII. References

[1] R.F. Cancho, et al., "Patterns in syntactic dependency networks", Physical Review, E 69, 051915, 2004.

[2] H. Liu and C. Hu, "Can syntactic networks indicate morphological complexity of a language", *EPL* 93 28005, 2001.

[3] J. Borge-Holthoefer and A. Arenas, "Semantic Networks: Structure and Dynamics", Entropy, 12, 1264-1302, 2010.

[4] S. Arbesman et al., "Comparative Analysis of Networks of Phonologically Similar Words in English and Spanish", Entropy, 12, pages 327-337, 2010.

[5] M. Medeiros Soares et al., "The Network of syllables in Portuguese", Physica A, 355(2-4): 678-684, 2005.

[6] K. Ban et al., "A preliminary study of Croatian Language Syllable Networks", Mipro SP, pp 1697-1701, 2013.

[7] K. Ban et al., "Initial comparison of linguistic networks measures for parallel texts", 5th International Conference on Information Technologies and Information Society (ITIS), pp. 97-104, 2013.

[8] D. Margan, S. Martinčić-Ipšić and A. Meštrović, "Preliminary report on the structure of Croatian linguistic co-occurrence networks", 5th International Conference on Information Technologies and Information Society, Slovenia, pp. 89-96, 2013.

[9] D. Margan, S. Martinčić-Ipšić and A. Meštrović, "Network Differences Between Normal and Shuffled Texts: Case of Croatian", Accepted at CompleNet 2014 Workshop, 2014.

[10] A. P. Masucci and G. J. Rodgers, "Differences between normal and shuffled texts: structural properties of weighted networks", Advances in Complex Systems, 12(01):113-129, 2009.

[11] R. F. Cancho and Richard V Solé. "The small world of human language", Proceedings of the Royal Society of London, 268(1482):2261–2265, 2001.

[12] S. N. Dorogovtsev and J. F. Mendes. "Language as an evolving word web", Proceedings of the Royal Society of London. Series B: Biological Sciences, 268(1485):2603–2606, 2001.

[13] A. P. Masucci and G. J. Rodgers. "Network properties of written human language", Physical Review E, 74(2):026102, 2006.

[14] H. T. Liu and J. Cong. Language clustering with word co-occurrence networks based on parallel texts. Chinese Science Bulletin, 58(10):1139–1144, 2013.

[15] D. R. Amancio et al. "Complex networks analysis of language complexity", *arXiv preprint arXiv:1302.4490* (2013).

[16] L. Antiqueira et al. "A complex network approach to text summarization", Information Sciences, 179(5), 2009, pp. 584-599

[17] D. R. Amancio et al. "Using metrics from complex networks to evaluate machine translation", *Physica A: Statistical Mechanics and its Applications*, *390*(1), pp. 131-142. 2009.

[18] L. Antiqueira et al. "Some issues on complex networks for author characterization." Inteligencia Artificial, Revista Iberoamericana de Inteligencia Artificial 11.36 (2007): 51-58.

[19] L. Antiqueira et al. "Strong correlations between text quality and complex networks features." *Physica A: Statistical Mechanics and its Applications* 373 (2007): 811-820.

[20] R.V. Solé, B.C. Murtra, S. Valverde, L. Steels. "Language Networks: their structure, function and evolution", Trends in Cognitive Sciences, 2005.

[21] M. Choudhury et al. "Global topology of word co-occurrence networks: Beyond the two-regime power-law", In Proceedings of International Conference on Computational Linguistics, pp 162–170, 2010.

[22] L.F. Costa, et. al. Boas, "Characterization of complex networks: a survey of measurements", cond-mat/ 0505185, 2005.

[23] Hagberget. Et al. "Exploring network structure, dynamics, and function using networkx". Technical report, (LANL), 2008.